\documentclass{article}
\usepackage{ijcai25}

\usepackage{times}
\usepackage{soul}
\usepackage{url}
\usepackage[hidelinks]{hyperref}
\usepackage[utf8]{inputenc}
\usepackage[small]{caption}
\usepackage{graphicx}
\usepackage{amsmath}
\usepackage{amsthm}
\usepackage{booktabs}
\usepackage{algorithm}
\usepackage[switch]{lineno}

\usepackage{multirow}
\usepackage{tabularx}
\usepackage[table]{xcolor}
\usepackage{subfig}
\usepackage{algpseudocode}
\usepackage{amssymb}
\usepackage{stmaryrd}
\usepackage{float}
\usepackage{xr}

\newcommand{\tablestyle}[2]{\setlength{\tabcolsep}{#1}\renewcommand{\arraystretch}{#2}\centering\footnotesize}
\newlength\savewidth\newcommand\shline{\noalign{\global\savewidth\arrayrulewidth
  \global\arrayrulewidth 1pt}\hline\noalign{\global\arrayrulewidth\savewidth}}

\newcolumntype{x}[1]{>{\centering\arraybackslash}p{#1pt}}
\newcolumntype{y}[1]{>{\raggedright\arraybackslash}p{#1pt}}
\newcolumntype{z}[1]{>{\raggedleft\arraybackslash}p{#1pt}}

\definecolor{baselinecolor}{gray}{.9}

\title{Why the Agent Made that Decision: Contrastive Explanation Learning for Reinforcement Learning}

\author{
Rui Zuo$^1$
\and
Simon Khan$^2$\and
Zifan Wang$^1$\and
Garrett Ethan Katz$^1$\And
Qinru Qiu$^1$ \\
\affiliations
$^1$Syracuse University\\
$^2$Air Force Research Laboratory\\
\emails
\{rzuo02, zwang345, gkatz01, qiqiu\}@syr.com,
simon.khan@us.af.mil
}

\begin{document}

\maketitle

\begin{abstract}
Reinforcement learning (RL) has demonstrated remarkable success in solving complex decision-making problems, yet its adoption in critical domains is hindered by the lack of interpretability in its decision-making processes. Existing explainable AI (xAI) approaches often fail to provide meaningful explanations for RL agents, particularly because they overlook the contrastive nature of human reasoning—answering "why this action instead of that one?" To address this gap, we propose a novel framework of contrastive learning to explain RL selected actions, named \textbf{VisionMask}. VisionMask is trained to generate explanations by explicitly contrasting the agent's chosen action with alternative actions in a given state using a self-supervised manner.
We demonstrate the efficacy of our method through experiments across diverse RL environments, evaluating it in terms of faithfulness, robustness and complexity. Our results show that VisionMask significantly improves human understanding of agent behavior while maintaining accuracy and fidelity. Furthermore, we present examples illustrating how VisionMask can be used for counterfactual analysis. This work bridges the gap between RL and xAI, paving the way for safer and more interpretable RL systems.
\end{abstract}

\section{Introduction}

Deep Reinforcement Learning (DRL) is a powerful technology in machine intelligence, widely used for many applications~\cite{sutton1999reinforcement}.
However, understanding a DRL agent's decision-making process is challenging, due to the inherent lack of explainability in the high-dimensional, non-linear structure of its underlying Deep Neural Network (DNN)~\cite{rl_lack_exp1,rl_lack_exp2}.
The lack of transparency undermines users' trust, driving the development of Explainable AI (xAI).

Various methods in computer vision have been proposed to enhance the transparency of AI systems \cite{LIME,gradcam,lrp,deeplift,SHAP}.
At the core, they share a common foundation: \textbf{attributing} the classifier's outputs to more interpretable features and using a saliency map to visualize these attributions. Their only differences are how these attributions are calculated.  A high-quality attribution-based explanation should meet several key criteria. First, it should demonstrate \emph{faithfulness}, meaning that including features with high attribution should lead the model to the target output, and excluding them should prevent it. Second, it should exhibit \emph{specificity}, ensuring that only critical features receive high attribution. Sometime this is also referred to as \emph{sparseness}. Finally, it should be \emph{robust}, meaning the explanation should remain consistent and not change significantly with minor variations in the input.

Attribution-based explanation has been studied for DRL models. %
\cite{pmlr-v80-greydanus18a} and %
~\cite{PuriVGKDK020} utilized policy distributional shifts as the basis for attribution in RL.
Specifically, they calculate attribution of a feature as the difference in Q/V values or action distributions between the original and perturbed states.
For example, given agent policy $\pi$, the attribution of a feature is proportional to $E_{s^\prime}(|\pi(s) - \pi(s^\prime)|_2)$ where $s$ stands for the original state and $s^\prime$ represents perturbed states  generated for this feature. By calculating the attributions for all features, a saliency map $m$ can be created.
However, the perturbation-based explanations lack faithfulness. Since each perturbation focuses only on local features while ignoring the joint impact of feature combinations, overlaying the saliency map with the original state, $(m \odot s)$, does not result in a feature combination that leads the agent to the target action distribution $\pi(s)$. %

A better approach to enhance faithfulness  is to \textbf{learn} a model to predict the saliency map $m$ that minimizes the difference between $\pi(m \odot s)$ and $\pi(s)$.
\textit{Explainer}~\cite{stalder2022you} leveraged this idea by training an explanation model for an image classifier. %
However, Explainer categorizes class labels into target and non-target for each training sample and focus on learning saliency map (or mask) only for the target label while treat all non-target labels as a single group. Unlike a (well trained) image classifier, where predictions for non-target labels are typically close to 0, DRL agent in many scenarios, does not exhibit a clear preference for the actions. Non-target actions may sometimes have probabilities only slightly lower than those of target actions. Analyzing how masking the feature may affect the non-target action probability provides additional information that can be used to train the explanation model more effectively. %

The above analysis motivates us to design a trainable saliency map generator for attribution-based explanations and train it using two channels of contrastive information:\textbf{(i) Action-wise contrast:}
We believe that environment states contain features that motivate the DRL agent to select both target action and non-target actions. However, the target action is ultimately chosen because it corresponds to higher reward or has a stronger presence. For each action, a saliency map can be generated as an explanation. 
Choosing features according to the saliency map for a non-target action should push the agent away from the target action, and vice versa.
This inspired us to treat the saliency map of the non-target action $m_{a'}$ and the target action $m_a$ as a negative pair, which can be leveraged for contrastive learning~\cite{contrative1,contrastive2,contrastive3}.
\textbf{(ii) Feature-wise contrast:}
To exclude irrelevant features (e.g. background) from the saliency map, explanations also need to be discriminative in filtering out such information.
When only irrelevant features are accessible to the agent, the resulting action distribution should be as uniform as possible.
Therefore, the target action's saliency map ($m$) and its inverted counterpart ($\Tilde{m} = 1 - m$) form another negative pair for contrastive learning.

In this work, we present VisionMask as an RL explainer that is contrastively trained to generate saliency maps to explain agent's actions.
We specifically focus on agents that maps images to actions and consider each pixel value as the interpretable input feature, although the similar technique could be extended to other type of features.
We carefully design the objective function to enable self-supervised contrastive learning of explanations from both action-wise and feature-wise perspectives, fostering the generation of more faithful explanations.
We conduct evaluation on six RL environments with five baselines based on faithfulness, robustness, and sparseness.
Quantitatively, VisionMask outperforms the baselines in terms of faithfulness, while exhibiting strong robustness and high sparseness.
Qualitatively, we compared VisionMask with the baselines in two settings: visual comparison and human studies.
In the visual comparison, VisionMask provides sharper explanations that align more closely with human interpretations, as demonstrated by counterfactual examples.
In the human studies, VisionMask's explanations help users better understand the agent's decisions and calibrate appropriate trust.

The contribution of our work can be summarized as the following:
\begin{itemize}
  \item We present VisionMask, a novel attribution method for explainable visual reinforcement learning that generates action-specific saliency maps.
  To the best of our knowledge, this is the first work that learns saliency map in a self-supervised contrastive learning manner, yielding highly faithful explanations across various RL environments.
  \item VisionMask is model-agnostic in that it works with any vision-based DRL agent and requires only the agent's input state and output action during inference time. VisionMask does not make any modification on the agent, however, each VisionMask model is trained specifically for a given agent by including the fixed agent in the back-propagation path for gradient descent. 
  \item We conducted extensive experiments, both qualitatively and quantitatively, to demonstrate the improvements of the saliency map in terms of faithfulness, robustness, and sparseness, as well as a comprehensive ablation study for each component.
\end{itemize}

\section{Background and Related Work}

The  nested structure and non-linear operation of DNN make it challenging for humans to understand how the outputs are derived from inputs. Some existing works address this challenge by explaining model outputs through input attributions \cite{petsiuk2018rise,LIME,DBLP:conf/iccv/FongV17}.
For example, Randomized Input Sampling for Explanation of Black-box Models (RISE) \cite{petsiuk2018rise} is a perturbation-based approach that explains neural image classifiers by  applying randomly generated masks to the input image and assessing their impact.
Local Interpretable Model-agnostic Explanations (LIME)\cite{LIME} tries to explain local instances by approximating them within a nearby vicinity using a linear decision model, where the explainable elements are superpixels (i.e., small image regions of similar pixels).

Built on top of the DNNs, the lack of transparency of the DRL agents undermines humans trust and hinders their adoption. 
Several studies have addressed this challenging problems.
According to \cite{qing2022survey}, existing efforts can be categorized into four main approaches: model explaining, reward explaining, task explaining, and state explaining.

Model explaining relies on inherently interpretable model architectures or auxiliary reasoning mechanisms to generate explanations \cite{topin2021iterative}.
However, these self-explanatory or symbolic models often suffer from decreased performance compared to state-of-the-art neural network based RL policies and may lack the representational power needed to learn more complex policies.
Reward explanation typically involves using explicitly designed explainable reward functions to generate explanations \cite{ashwood2022dynamic}.
Task explanation considers a policy as a composition of sub-task and action sequences, explaining the behavior in terms of relationships among sub-tasks \cite{shu2017hierarchical}.
These approaches often assume the existence of a reward decomposition scheme or a predefined sub-task partition mechanism, which may not hold true in all RL environments.

State explanations are based on agent observations  from the environment.
This approach determines the significance of \emph{explainable elements} within a state observation in relation to action selection or reward determination. An explainable element in the state could be a small region in the visual input or semantic features in the environment state.
The proposed \emph{VisionMask} falls into this category.

State explaining methods can further be divided into three categories: attention-based, Shapley value-based and perturbation-based mechanisms, which are detailed below. 
Attention is a common approach used for explainable RL in several existing works \cite{annasamy2019towards}.
However, similar to i-DQN \cite{annasamy2019towards}, these methods cannot explain a given DRL model because their attention model must be trained alongside the agent model. 
Shapley value \cite{shapley1953value}, a concept from game theory, has also been introduced into XRL.
SVERL \cite{beechey2023explaining} provides a theoretical analysis on how Shapley values can be applied to explain value and policy network in DRL. However, this approach is still in its early stages and assumes prior knowledge of the transition model of the environment, which is not realistic for realistic applications.  %

Perturbation-based methods \cite{pmlr-v80-greydanus18a,iyer2018transparency,PuriVGKDK020} compute saliency maps of the input features by comparing the action probability or value function before and after perturbation.
\cite{pmlr-v80-greydanus18a} perturbed the state with Gaussian blur at fixed intervals,
while \cite{iyer2018transparency} identified objects and perturbed them with a constant gray value. \cite{PuriVGKDK020} improved on \cite{pmlr-v80-greydanus18a} by removing the effect of perturbations from other irrelevant actions.
As discussed in section 1, the performance of these techniques is constrained by predefined perturbation rules and the lack of knowledge accumulation. Furthermore, there is no guarantee that the perturbed input remains physically meaningful.

\begin{figure*}[t] \label{visionmask_arch}
    \centering
    \includegraphics[width=0.9\linewidth]{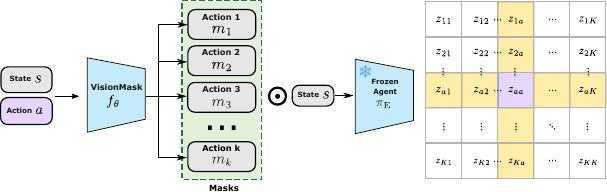}
    \caption{\small\textbf{Architecture of VisionMask.} State-action pairs $(s, a)$ are sampled from $\mathcal{D}_E$.}
    \label{fig:archi} 
\end{figure*}

\section{VisionMask}
In this section, we present our VisionMask architecture. The primary goal is to generate action-wise saliency maps that attribute the most relevant features in the state to each action. For agents that map images to actions, the features and states correspond to pixels and images.

\subsection{Problem Formulation}
\label{sec:problem_formulation}

Formally, we define the environment as a Markov Decision Process (MDP) $\{ \mathcal{S}, \mathcal{A}, P, R, \gamma \}$, where $\mathcal{S}$ represents the state space; $\mathcal{A}$ denotes the action space with $|\mathcal{A}|=K$; the state transition function $P: \mathcal{S} \times \mathcal{A} \to \Delta(\mathcal{S})$ depicts the transition between states based on actions, where $\Delta(\mathcal{S})$ represents the set of probability distributions over states; the reward function $R: \mathcal{S} \times \mathcal{A} \to \mathbb{R}$ provides the immediate reward for state-action pairs; $\gamma \in [0,1]$ is discount factor and $\pi: \mathcal{S} \to \Delta(\mathcal{A})$ represent policy.
Return $G$ is defined as $G = \sum_{k=0}^\infty\gamma^kR_{k+1}$,
and the expected cumulative reward of a policy $\pi$ is
$
\mathbb{E}_\pi[G] = \mathbb{E}_\pi[\sum_{k=0}^\infty\gamma^kR_{k+1}]
$, where the expectation is taken with respect to the initial state distribution, transition probabilities, and action probabilities determined by $\pi$.
VisionMask operates on a given trained expert policy $\pi_{E}$ such that
$
\pi_E \approx \pi^{*} = \arg\max_{\pi}\mathbb{E}_{\pi}[G],
$
where $\pi^*$ is the optimal policy.
We can obtain a dataset of expert demonstrations $\mathcal{D}_E = \{(s_i, a_i \sim \pi_E(s_i))\}_{i=1}^N$ consisting of $N$ state-action pairs, from trajectories sampled while executing $\pi_E$ in the environment.
Our goal is to learn an explainer $f_{\theta^*}$ that minimizes the loss
$
\theta^* = \arg\min_{\theta} \sum_{(s, a) \in \mathcal{D}_E} \mathcal{L} (a, s, \theta)
$
where $\mathcal{L}$ is the training loss function to be discussed in section 3.3. The explainer function 
$f_{\theta}: \mathcal{S} \to [0, 1]^{K \times d_s}$, where $d_s$ represents the feature size of the state $s\in\mathcal{S}$ and $K$ denotes the number of candidate actions, predicts the attributions of each action to each feature in the state $s$. 
The value of the output is bounded within the range $[0, 1]$, with the $(i,j)$th element indicating the $j$th feature's attribution to the $i$th action.

In the case of visual DRL, $s$ is the visual input of the agent and $d_s$ is the number of pixels in $s$, i.e., $d_s=W\times{H}$, with $W$ and $H$ representing the width and height of the visual input. The output of the explainer can be viewed as $K$ saliency maps, $M = \{m_0, m_1, ...., m_{K-1}\}$, each corresponds to an action.  A saliency value $m_i[x,y]$ near 1 indicates that the pixel $(x,y)$ has significant contributions to action $i$, whereas a value near 0 denotes irrelevance. Overlaying the $i$th saliency map with the state $s$ will highlight the input features that lead the agent to the $i$th action.

\subsection{Architecture}

As shown in Figure~\ref{fig:archi}, we first collect the expert dataset $\mathcal{D}_E$ using the expert policy $\pi_E$. From this dataset $\mathcal{D}_E$, state-action pairs $(s_i, a_i)$ are sampled and fed to VisionMas $f_\theta$ to generate the set of saliency maps $M$.

Generating the saliency map from given visual input is a dense prediction task that shares similarities with image segmentation, where each pixel is assigned a value to indicate whether it belongs to an object or background.
Hence, we structure the explainer $f_\theta$ akin to the widely used image segmentation model,
DeepLabv3 \cite{deeplabv3}, however, retrain it using self-supervised contrastive learning.
To make sure that the output saliency value are bounded to the range $[0, 1]$, a sigmoid function is applied at the output of $f_\theta$. 
For each $m_i \in M$, we also calculate a complement map $\Tilde{m}_i = \mathbf{1} - m_i$ highlighting the irrelevant regions for the action $i$.
Then the masks $m_i$ and $\Tilde{m}_i$ are overlaid onto the original state $s$ to generate two masked states $s_i$ and $\Tilde{s}_i$ using the following overlay function:
\begin{equation} \label{eq1}
s_{i} = s \odot m_i + r \odot (\Tilde{m}_i), 
\Tilde{s}_i = s \odot \Tilde{m}_i + r \odot (m_i)
\end{equation}
where $\odot$ is Hadamard Product and $r$ is a reference value. 
Numerous options exist for the reference value, such as setting the pixel to zero, assigning a constant value, blurring the pixel, or cropping it.
Empirical study shows that setting the reference to the background gives the best results. More details can be found in the ablation study~\ref{sec:ablation}. 
See Appendix \ref{app:sec_ref_value} for the background values for different environments.

To generate self-supervised contrastive loss to train the model, we query the agent to obtain the corresponding logits $z_i = z(s_{i})$ and $\Tilde{z}_i = z(\Tilde{s}_i)$, where $z_i, \tilde{z}_i \in \mathbb{R}^K$ , and the action probability distributions $p_i = \text{softmax}(z_i)$ and $\Tilde{p}_i = \text{softmax}(\Tilde{z}_i)$,  where $p, p_i \in [0, 1]^K$, $0\leq{i}<K$.
By concatenating each $p_i$ and $\Tilde{p}_i$, we have the the action probability distributions of each mask $\mathbf{p}, \mathbf{\Tilde{p}} \in \mathbb{R}^{K \times K}$,
\begingroup
\setlength{\abovedisplayskip}{5pt}
\setlength{\belowdisplayskip}{5pt}
\begin{equation*}
   \mathbf{p} = [p_1, p_2, \dots, p_k]^T \quad \mathbf{\Tilde{p}} = [\Tilde{p}_1, \Tilde{p}_2, \dots, \Tilde{p}_k]^T
\end{equation*}

\subsection{Training Loss}
To enable the agent contrastively learn the saliency map $m_a$, we carefully designed the training loss function $\mathcal{L}$ as follows:
\begin{equation*}
\mathcal{L}(\mathbf{s}, \mathbf{a}, \theta) = \mathcal{L}_{a}(\mathbf{s}, \mathbf{a}) + \lambda_{ne} \mathcal{L}_{ne}(\mathbf{s})
+ \lambda_{area} \mathcal{L}_{area}(\mathbf{s}, \mathbf{a})
\end{equation*}
where $\mathcal{L}_{a}(\mathbf{s}, \mathbf{a})$ is action-wise contrastive action loss, $\mathcal{L}_{ne}(\mathbf{s}, \mathbf{a})$ is feature-wise loss, $\mathcal{L}_{area}(\mathbf{s}, \mathbf{a}, \mathbf{n})$ is the area size loss  and $\lambda_{ne}, \lambda_{area}$  are regularization hyper-parameters.

\textbf{Action-wise contrastive loss} $\mathcal{L}_{a}$.
Let $a$ denote the target action chosen by the agent. Our primary goal is to learn explanations faithful to the  $a$, making $(a, p_a)$ the only positive pair.
Furthermore, the explanation must be discriminative, meaning it should clearly distinguish the target action $a$ from all other possible actions. As a result, every other pair of $(a, p_{i \neq a})$ is treated as a negative pair.
Then, we compute the cross-entropy loss between these pairs,
\begin{align}
\mathcal{L}_{pos}(\mathbf{s}, \mathbf{a}) = - \frac{1}{K} \sum_{k=1}^{K} \llbracket k = \mathbf{a} \rrbracket \log(\frac{\text{exp}(z_{aa})}{\sum_{i=1}^{K}\text{exp}(z_{ai})})\label{loss:positive} \\
\mathcal{L}_{neg}(\mathbf{s}, \mathbf{a}) = - \frac{1}{K} \sum_{k=1}^{K} \llbracket k = \mathbf{a} \rrbracket \log(\frac{\text{exp}(z_{aa})}{\sum^{K}_{i=1}\text{exp}(z_{ia})})
\end{align}
where $z_{ij}$ is the logit for action $j$ when querying the policy with the masked state $s_i$. 
The contrastive action loss $\mathcal{L}_a = \mathcal{L}_{pos} + \mathcal{L}_{neg}$.
Here $\llbracket k = \mathbf{a} \rrbracket$ denotes the indicator function which returns 1 if $k$ is the same as label $\mathbf{a}$, and 0 otherwise.
Note that we do not compute the loss with $\mathbf{p}_{k,i \neq a}$, as ensuring the faithfulness of the target action $a$ is our primary objective here.

\textbf{Feature-wise loss} $\mathcal{L}_{ne}$.
To ensure that the visual input regions selected by $m_i$ is necessary for the agent to make decisions, we also need to make sure that the unselected region, i.e., $s_{\Tilde{m_i}}$, does not provide useful information for action selection, hence the action distribution $\mathbf{\Tilde{p}}$ should follow a uniform distribution. Motivated by this rationale, we define negative entropy loss regarding $\Tilde{\mathbf{m}}$ as the following:
\begin{align}
    \mathcal{L}_{ne}(\mathbf{s}) = \frac{1}{K^2}\sum_{ij} \Tilde{\mathbf{p}}_{ij} \log \Tilde{\mathbf{p}}_{ij}.
\end{align}

\textbf{Area size loss $\mathcal{L}_{area}$}.
A low effort way to minimize  $\mathcal{L}_{a}$ and $\mathcal{L}_{ne}$ is to include all pixels in the mask, $m_i$, and no pixel in the complement mask, $\Tilde{m_i}$, which obviously is not a valid solution. We need to ensure that each importance mask only consists a small number of crucial pixels. Thus, we define $\mathcal{L}_{area}$ using L1 norm as follows:
\begin{align}
    \mathcal{L}_{area}(\mathbf{s}) = \frac{1}{K}\sum_{k}(|\frac{1}{Z}\sum_{i,j}m_k[i,j] - a_{max}|)
\end{align}
where $Z$ it the number of pixels in state.

\begin{table*}[ht]
\centering
\begin{minipage}{\linewidth}{\begin{center}
\tablestyle{4pt}{1.05}
\begin{tabular}{l|lllll|lllll|llllll}
 & \multicolumn{5}{c|}{\textit{SMB}} & \multicolumn{5}{c|}{\textit{Enduro}} & \multicolumn{5}{c}{\textit{Seaquest}}\\
Method & Acc. &  Del. $\downarrow$ & Ins. $\uparrow$ & LLE $\downarrow$ & Sp. $\uparrow$ & Acc. &  Del. $\downarrow$ & Ins. $\uparrow$ & LLE $\downarrow$ & Sp. $\uparrow$ & Acc. &  Del. $\downarrow$ & Ins. $\uparrow$ & LLE $\downarrow$ & Sp. $\uparrow$ \\
\shline
LIME &
78.7 & 27.0 & 30.2 & 69.4 & \textbf{98.3} &
88.1 & 36.3 & 38.0 & 47.9 & \textbf{90.2} & 
90.1 & 14.3 & 21.4 & 17.3 & \textbf{97.7} \\
RISE &
80.5 & \textbf{18.9} & 38.0 & \textbf{1.9} & 2.0 &
90.3 & 37.6 & 39.1 & \textbf{0.01} & 0.5 & 
92.0 & 8.1 & 29.0 & \textbf{0.02} & 1.4 \\
Greydanus &
16.9 & 56.4 & 20.2 & \textcolor{blue}{20.8} & 63.3 &
27.4 & 34.6 & 33.3 & 0.25 & \textcolor{blue}{82.9} &
44.6 & 14.4 & 7.88 & \textcolor{blue}{0.12} & 75.1 \\
SARFA &
16.9 & 55.6 & 21.4 & 68.7 & 65.5 &
27.9 & 33.3 & 33.4 & 0.26 & 77.0 &
44.8 & 7.6 & 8.7 & 0.13 & 75.2  \\
\textit{Explainer} &
92.2 & 23.6 & 49.0 & 38.1 & 81.2 &
90.2 & 34.2 & 33.1 & 0.22 & 81.2 &
95.3 & 13.4 & 30.0 & 0.9 & 97.5 \\
VisionMask &
\textbf{95.9} & 20.4 & \textbf{67.6} & 38.0 & \textcolor{blue}{82.3} & 
\textbf{98.7} & \textbf{32.9} & \textbf{41.2} & \textcolor{blue}{0.5} & 80.0 &
\textbf{99.6} & \textbf{6.4} & \textbf{34.3} & 0.9 & \textcolor{blue}{97.6} \\
 & \multicolumn{5}{c}{\textit{MsPacman}} & \multicolumn{5}{c}{\textit{VizDoom}} & \multicolumn{5}{c}{\textit{Highway}}\\
Method & Acc. &  Del. $\downarrow$ & Ins. $\uparrow$ & LLE $\downarrow$ & Sp. $\uparrow$ & Acc. &  Del. $\downarrow$ & Ins. $\uparrow$ & LLE $\downarrow$ & Sp. $\uparrow$ & Acc. &  Del. $\downarrow$ & Ins. $\uparrow$ & LLE $\downarrow$ & Sp. $\uparrow$ \\
\shline
LIME &
84.5 & 32.7 & 37.7 & 9.1 & \textbf{96.6} &
72.1 & 35.1 & 11.2 & 22.1 & \textbf{96.4} &
77.8 & 20.0 & 23.64 & 52.7 & 80.5\\
RISE &
92.6 & 26.2 & 45.7 & \textbf{0.01} & 0.59 &
83.8 & 14.6 & 14.5 & \textbf{0.02} & 0.5 &
82.5 & 20.0 & 22.1 & \textbf{0.05} & 0.6 \\
Greydanus &
46.1 & 39.1 & 14.2 & 0.17 & 57.4 &
75.2 & 20.7 & 17.2 & \textcolor{blue}{0.12} & 62.1 &
92.7 & 20.9 & 24.7 & 0.93 & 83.8 \\
SARFA &
42.4 & 18.8 & 15.6 & \textcolor{blue}{0.13} & 66.5 &
76.1 & 16.7 & 17.4 & 0.55 & \textcolor{blue}{66.6} &
97.6 & 20.4 & 25.6 & \textcolor{blue}{0.43} & 83.0 \\
\textit{Explainer} &
97.6 & 19.6 & 59.2 & 0.22 & 81.2 &
84.2 & 14.4 & 17.6 & 0.2 & 65.5 &
95.0 & 20.37 & 24.4 & 1.5 & \textcolor{blue}{83.9} \\
VisionMask &
\textbf{98.7} & \textbf{17.0} & \textbf{62.8} & 0.2 & \textcolor{blue}{89.5} &
\textbf{87.8} & \textbf{14.2} & \textbf{18.1} & 3.89 & 47.8 &
\textbf{98.1} & \textbf{19.8} & \textbf{25.8} & 1.14 & \textbf{84.8} \\
\end{tabular}
\end{center}}\end{minipage}
\caption{\textbf{Quantitative results} on \textit{SMB}, \textit{Enduro}, \textit{Seaquest}, \textit{MsPacman}, \textit{VizDoom} and \textit{Highway} of VisionMask against 5 baselines. Five metrics are compared. The faithfulness is measured by Accuracy(Acc.), Deletion(Del.) and Insertion(Ins.) metrics(\%); Robustness is measured by Local Lipschitz Estimate (LLE)(\%); And Complexity is measured by Sparseness(Sp.)(\%). \textcolor{blue}{Blue} represents second best results.
}
\label{tab:quant} 
\end{table*}

\begin{figure*}[ht]
\centering
\captionsetup[subfloat]{labelformat=empty,font=footnotesize}
\captionsetup[table*]{font=footnotesize}
\subfloat[
]{
\centering
\begin{minipage}{\linewidth}{\begin{center}
\includegraphics[width=0.9\linewidth]{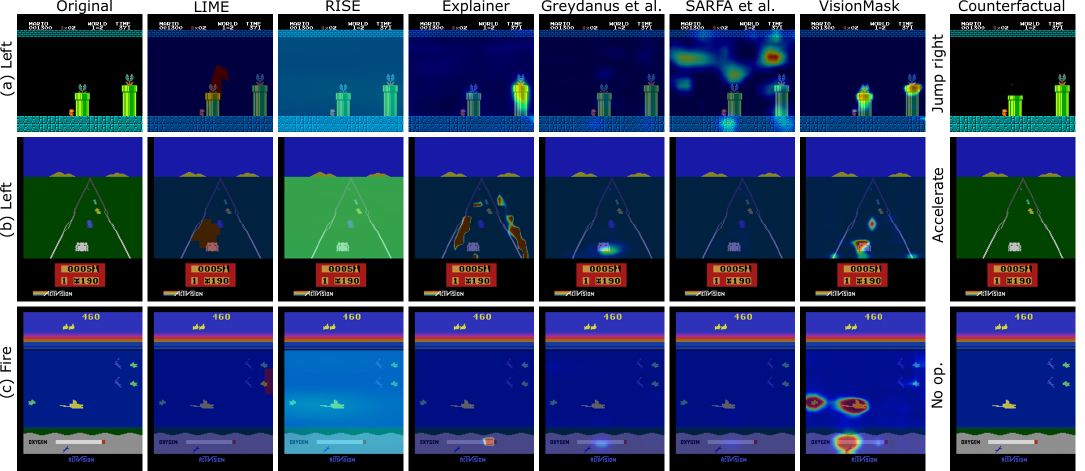}
\end{center}}\end{minipage}
}
\caption{
\textbf{Qualitative examples} of VisionMask and five baselines across three environments.
(a)–(c) show the saliency map overlaid on the input image and counterfactuals where regions are removed based on VisionMask's map.
(a) Human: "Mario moves left to avoid the Piranha Plant." VisionMask correctly highlights the Plant; removing it changes the action from 'Left' to 'Jump right'.
(b) Human: "The agent moves left at constant speed due to the front car." VisionMask identifies both; removing the car allows acceleration instead of 'Right + accelerate'.
(c) Human: "The agent fires because a shark follows." VisionMask detects the shark and oxygen bar; removing the shark stops firing.
Additional examples in Appendix~\ref{app:additional_quali}.
}
\label{fig:quali_and_cf} 
\end{figure*}

\section{Experiments} \label{sec_exp}

In this section, we begin by outlining the experimental setup. We then present quantitative and qualitative analyses to evaluate our approach. Additionally, we provide counterfactual explanations to demo VisionMask's faithfulness and sensitivity. Finally, we perform an ablation study to assess the contribution of each component.

\subsection{Experimental Setup}
\textbf{Environment Selection}. We conduct experiments across three types of environments: \textit{Super Mario Bros (SMB)} \cite{gym-super-mario-bros}, \textit{Enduro}, \textit{Seaquest} and \textit{MsPacman} \cite{atari1,atari2} for 2D game, \textit{Highway-env} \cite{highway-env} for autonomous driving simulation and \textit{VizDoom} \cite{vizdoom} for 3D game.

\textbf{Baseline Selection}.
We mainly compare our model with perturbation-based baselines for black-box RL such as Greydanus \cite{pmlr-v80-greydanus18a} and \textit{SARFA} \cite{PuriVGKDK020}.
In addition, we also compared with three techniques originally designed to explain image classifiers, including a learning-based method, \textit{Explainer}~\cite{stalder2022you}, and two perturbation-based methods, LIME \cite{LIME} and RISE \cite{petsiuk2018rise}. %
Although these methods focus on image classification, their main concept is similar to ours: %
to attribute a given label (action) to a subset of visual features.
We use the public implementation from torchray \cite{torchray} for RISE and the original published implementations for other baselines. Among all the baselines, the Explainer is the most similar to VisionMask, as both are learning-based approaches. However, VisionMask employs action-contrastive learning and is trained using different regularization. In the experimental results, we demonstrate that these differences significantly enhances VisionMask's performance.

\textbf{Expert Policy}.
We use the open-source pre-trained PPO \cite{schulman2017proximal} agents from \cite{super-mario-bros-PPO-pytorch} for \textit{SMB} environment and from stable-baselines3 \cite{stable-baselines3} for \textit{Enduro}, \textit{Seaquest} and \textit{MsPacman} environments.
We train the DQN \cite{DQN} agents from scratch for the \textit{VizDoom} and \textit{Highway-env} environments.
See Appendix~\ref{app:agent} for agents details.

\textbf{Dataset.}
We collect state-action pairs for each environment and split the data into 80\% for training, 10\% for validation, and 10\% for testing. All results in this section are reported using the test split.
We make this dataset publicly available; see Appendix~\ref{app:dataset} for details.

\subsection{Quantitative Analysis} \label{sub:quant}

\textbf{Metrics.} Since there is no ground truth explanations~\cite{no_ground_truth}, it is crucial to select appropriate metrics to evaluate the trustworthiness of explanations.
Designing robust and consistent metrics for XAI remains an unresolved challenge~\cite{unresolved}. 
To alleviate the inconsistency between metrics and avoid selection bias, we closely follow the LATEC~\cite{LATEC} benchmark to evaluate performance across three dimensions: \textit{Faithfulness}, %
\textit{Robustness}, %
and \textit{Complexity}. %

Faithfulness measures the correlation between the agent's action and the masked visual input. In this context, we consider three metrics: Accuracy, Deletion and Insertion performance ~\cite{petsiuk2018rise}. The accuracy gives the percentage of time that masked input and the original input lead to the same action according to the expert agent. It is defined as the following: 
$
\text{Accuracy} = \frac{1}{|\mathcal{D}_{\text{test}}|} \sum_{i=1}^{|\mathcal{D}_{\text{test}}|} \llbracket \pi_E(s_i) = \pi_E(s_i^{\prime}) \rrbracket,
$
where $s_i^{\prime} = s_i \odot m_i + r \odot{\Tilde{m_i}}$, and $\llbracket\cdot\rrbracket$ again denotes the indicator function.  

Insertion and deletion measure the impact on the target action probability by progressively  adding or removing pixels from the original input based on the descending order of their attribution scores, with the most important pixels being inserted/deleted first. The inserted/deleted input image will be processed by the expert agent again to obtain the probability of the target action. By plotting
the probability against the number of pixels added or removed, we obtain an
insertion and a deletion curve. The insertion/deletion performance is measured by the area under the curve (AUC) of the insertion/deletion curves. A larger (smaller) AUC for the insertion (deletion) curve means that including (removing) the important pixels identified by the explainer can effectively increase (reduce) the probability of the target action. Hence a larger (smaller) AUC of insertion (deletion) curve indicates more accurate attribution prediction. %
The pseudo-code of the detailed information of deletion and insertion could be found in Appedix~\ref{app:alg_del}.

For robustness, we report the Local Lipschitz Estimate (LLE) scores~\cite{LLE}, which quantify the local smoothness of explanations by estimating the Lipschitz constant within a specific neighborhood. The Lipschitz constant measures the maximum rate of change of the function, ensuring explanation does not vary too rapidly within the state's neighborhood.
Given state $s_i$ and neighborhood size $\epsilon$, LLE defined as
$$\hat{L}(s_i) = \underset{s_j \in \mathcal{N}_\epsilon(s_i)}{\text{argmax}} 
\frac{\|f_{\theta}(s_i) - f_{\theta}(s_j)\|_2}{\|s_i - s_j\|_2}$$
where $\mathcal{N}_\epsilon(s) = \{s' \in X \mid \|s - s'\| \leq \epsilon\}$.

We evaluate complexity with Sparseness~\cite{sparseness}, which uses the Gini index on the vector of absolute attribution values sorted in non-descending order.
Sparseness ensures that features genuinely influencing the output have substantial contributions, while insignificant or only slightly relevant features should have minimal contributions.
A higher Sparseness indicates more contrastive attribution values and hence more understandable explanations  ~\cite{sparseness}.

\textbf{Results.}
In Table~\ref{tab:quant}, our VisionMask achieves the best performance in terms of faithfulness (i.e., Acc, Del, Ins) in all testing environment except \textit{SMB}, where its deletion score is slightly lower than RISE. However, RISE has significantly lower insertion score and accuracy in this environment. Hence, our explanations are more aligned with the agent's decision-making process.
Moreover, compared to, \textit{Explainer}~\cite{stalder2022you}, VisionMask exhibits much higher faithfulness, which suggests the effectiveness of action contrastive learning. %
Overall, learning-based model performs better compared to perturbation based approach.

In terms of robustness, RISE achieves the best performance in terms of LLE score across all settings. This is because, as a perturbation-based method, RISE uses the same type of perturbation that we used to generate the neighbor image for LLE score calculation. Hence the perturbation almost has no impact to it.  On the other hand, RISE has the lowest sparsity, which means the attribution predicted by RISE are evenly distributed and contains little information.
For Complexity, benefiting from the binary mask of LIME which is much more sparse compared to other baselines, LIME has the highest sparseness score. However, it has the worst performance in terms of LLE score. 
Overall, VisionMask achieves the best balance between the robustness and sparseness.
See Appendix for the Radar map.

\begin{table*}[h]
\centering
\subfloat[
\textbf{Area size regularization}. L1 is better for faithfulness and faster. Time is estimated by training 10 epochs.
\label{tab:ablation_area_reg}
]{
\centering
\begin{minipage}{0.29\linewidth}{\begin{center}
\tablestyle{4pt}{1.05}
\begin{tabular}{lc>{\columncolor[gray]{.9}}c}
Metrics & Stalder &  L1 \\
\shline
Acc. &
95.6 & \textbf{95.9} \\
Ins. $\uparrow$ &
64.3 & \textbf{67.6} \\
Del. $\downarrow$ &
21.1 & \textbf{20.4} \\
LLE  $\downarrow$ &
\textbf{56.6} & 62.6 \\
Sp.  $\uparrow$ &
\textbf{73.8} & 62.3 \\
Hours & 0.72 & \textbf{0.68} \\
Speedup & - & $1.1\times$\\
Sp./LLE &
\textbf{1.3} & 1.0 \\
\end{tabular}
\end{center}}\end{minipage}
}
\hspace{2em}
\subfloat[
\textbf{$f_\theta$ model}. DeepLabv3 achieves better results.
\label{tab:ablation_seg_model}
]{
\begin{minipage}{0.29\linewidth}{\begin{center}
\tablestyle{4pt}{1.05}
\begin{tabular}{lcc>{\columncolor[gray]{.9}}c}
Metrics & FCN & UNet & DeepLabv3 \\
\shline
Acc. &
94.2 & 94.7 & \textbf{96.0} \\
Ins. $\uparrow$ &
56.5 & 56.8 & \textbf{67.5} \\
Del. $\downarrow$ &
\textbf{19.6} & 19.7 & 20.4 \\
LLE  $\downarrow$ &
100.0 & 86.3 & \textbf{62.6} \\
Sp.  $\uparrow$ &
58.0 & 19.8 & \textbf{62.3} \\
Sp./LLE &
0.58 & 0.23 & \textbf{0.99} \\
\multicolumn{4}{c}{~}\\
\multicolumn{4}{c}{~}\\
\end{tabular}
\end{center}}\end{minipage}
}
\hspace{2em}
\subfloat[
\textbf{Max area size $a_\text{max}$}. Maximum area size of $0.1$ produces favorable results.
\label{tab:ablation_maxarea_size}
]{
\begin{minipage}{0.29\linewidth}{\begin{center}
\tablestyle{4pt}{1.05}
\begin{tabular}{l>{\columncolor[gray]{.9}}ccc}
Metrics & $0.1$ & $0.2$ & $0.4$ \\
\shline
Acc. &
\textbf{95.9} & 94.5 & 95.8 \\
Ins. $\uparrow$ &
\textbf{57.0} & 56.9 & \textbf{57.0} \\
Del. $\downarrow$ &
19.5 & 19.5 & 19.5 \\
LLE  $\downarrow$ &
96.0 & \textbf{93.8} & 99.5 \\
Sp.  $\uparrow$ &
\textbf{79.0} & 58.6 & 48.1 \\
Sp./LLE &
\textbf{0.82} & 0.62 & 0.48 \\
\multicolumn{4}{c}{~}\\
\multicolumn{4}{c}{~}\\
\end{tabular}
\end{center}}\end{minipage}
}
\\
\centering
\subfloat[
\textbf{Reference value}. Using background produce better faithfulness values. 
\label{tab:ablation_refv}
]{
\begin{minipage}{0.4\linewidth}{\begin{center}
\tablestyle{4pt}{1.05}
\begin{tabular}{l>{\columncolor[gray]{.9}}cccc}
Metrics & background & black & mean & blur \\
\shline
Acc. &
\textbf{97.8} & 95.9 & 97.6 & 94.4 \\
Ins. $\uparrow$ &
\textbf{67.5} & 51.7 & {54.5} & 56.9 \\
Del. $\downarrow$ &
20.4 & \textbf{16.8} & 18.2 & 25.0 \\
LLE  $\downarrow$ &
62.6 & \textbf{54.4} & 92.6 & 99.6 \\
Sp.  $\uparrow$ &
\textbf{62.3} & 28.5 & 45.5 & 56.6 \\
Sp./LLE &
\textbf{1.0} & 0.52 & 0.49 & 0.57 \\
\end{tabular}
\end{center}}\end{minipage}
}
\hspace{2em}
\subfloat[
\textbf{Component}. Negative Entropy (NE) with L1 area regularization performs better in terms of faithufulness. Contrastive-only (C.), Stalder Area Loss (S.), and Total Variation (TV.)
\label{tab:ablation_component}
]{
\centering
\begin{minipage}{0.45\linewidth}{\begin{center}
\tablestyle{4pt}{1.05}
\begin{tabular}{lcc>{\columncolor[gray]{.9}}cc}
Comp. & C. & +NE & +NE+L1 & +NE+L1+TV \\
\shline
Acc. &
\textbf{96.7} & 96.5 & 95.9 & 95.6 \\
Ins. $\uparrow$ &
56.7 & {56.5} & \textbf{67.6} & 64.6 \\
Del. $\downarrow$ &
23.7 & {23.8} & \textbf{20.4} & \textbf{20.7} \\
LLE  $\downarrow$ &
99.3 & {96.0} & 62.6 & 68.8 \\
Sp.  $\uparrow$ &
24.0 & 23.8 & 62.3 & 60.5 \\
Sp./LLE &
0.24 & 0.25 & \textbf{0.99} & 0.88 \\
\end{tabular}
\end{center}}\end{minipage}
}
\caption{\textbf{VisionMask ablation experiments} on \textit{SMB}. Default settings are marked in \colorbox{baselinecolor}{gray}.}
\label{tab:ablations} 
\end{table*}

\subsection{Qualitative Analysis} \label{sub:quali}
In this section, we conduct two qualitative analyses across two settings, visual comparision and human studies.

\textbf{Visual Comparision}. 
We present example explanations from three environments, \textit{SMB}, \textit{Enduro}, and \textit{Seaquest}, in Figure~\ref{fig:quali_and_cf}, along with some counterfactual analysis generated from the explanations provided by VisionMask.
The examples show that both LIME and RISE fail to generate interpretable explanations. LIME’s superpixels are too large to capture the specific regions, while RISE's explanations include almost all pixels.
\textit{Explainer} generates more accurate and interpretable explanations compared to LIME and RISE. This suggests that purely perturbation-based approaches may fail in RL due to the high dynamics of environments and the lack of learning ability. %
\cite{pmlr-v80-greydanus18a} and SARFA generate better explanations than LIME and RISE but often focus on irrelavent objects or background. %
In contrast, VisionMask accurately highlights the relevant regions, providing sharp explanations that are both accurate and interpretable. %
By removing some of the regions highlighted by the VisionMask, counterfactual analysis could be performed to answer questions like: "Why this action instead of that one?"

\begin{figure}[htbp]\centering
\includegraphics[width=0.92\linewidth]{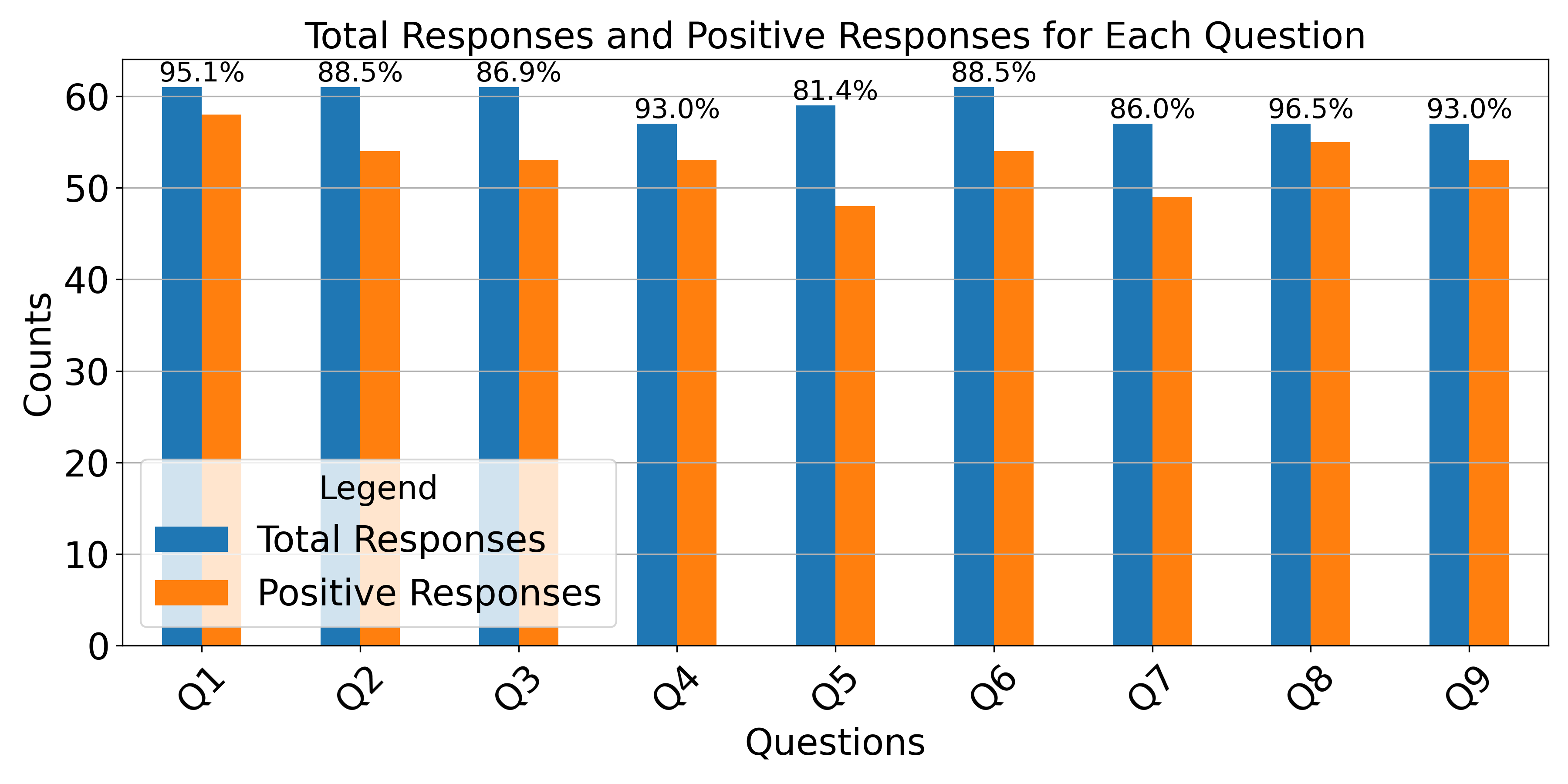}
\caption{\textbf{Human studies}. Comparison of total responses and positive responses across questions.}
\label{fig:survey}
\end{figure}

\textbf{Human Studies}. To assess whether the saliency maps improve humans' understanding of the agent's decisions, we conduct human studies for \textit{SMB}, \textit{Enduro} and \textit{Seaquest}.
We present 63 participants with 9 state-action pairs accompanied by the saliency maps generated by VisionMask. Participants are then asked whether the saliency maps help them better understand the decision-making process of the RL agents. We record the total number of responses as well as the number of positive responses, as shown in Figure~\ref{fig:survey}. Over 89\% of participants find the saliency maps generated by VisionMask helpful in understanding the agents' decisions. Details about this study can be found in the Appendix~\ref{app:human_study}.

\subsection{Ablation Studies}\label{sec:ablation}
Table~\ref{tab:ablations} shows the results for ablation studies. In all tables, the column marked in \colorbox{baselinecolor}{gray} represents the default setting of VisionMask used to generate results in previous sections.

\textbf{Area size regularization}. 
In this experiment, we replace the default L1 area regularization with the \textit{Min-Max Area Loss} from
\textit{Explainer}~\cite{stalder2022you}. Without this constraint, \textit{Explainer} tends to produce masks where all values are one in order to achieve the best performance.
\textit{Explainer} first vectorizes and sorts the mask elements, and then applies penalties to pixels falling outside the range $[a, b]$, where the hyperparameters $a$ and $b$ specify the minimum and maximum allowable area sizes.
Instead, VisionMask uses L1 area regularization to constrain the maximal area, as we do not wish to impose constraints on the minimum area size.
To implement action contrastive learning, we need to generate masks for actions that are not selected, and these masks should be allowed to be all zeros if necessary.
As shown in Table~\ref{tab:ablation_area_reg}, using L1 area regularization improves the faithfulness of the explanation and increases training speed.

\textbf{Max area size}.
We further vary the maximum permitted area size $a_{\text{max}}$ to evaluate its impact. As shown in Table~\ref{tab:ablation_maxarea_size}, reducing $a_\text{max}$ significantly increases the sparseness as it constrains the model to focus on the most discriminative regions without compromising other performance metrics.
The $0.1$ setting achieves the best results, as expected.

$\mathbf{f_\theta}$ \textbf{model}.
In Table~\ref{tab:ablation_seg_model}, we compare the performance of three different versions of VisionMask with three different segmentation models: our default DeepLabv3~\cite{deeplabv3}, Fully Convolutional Network (FCN)\cite{FCN}, and UNet\cite{UNet}. The results show that DeepLabv3 achieves a better performance.

\textbf{Reference Value.}
We evaluate the impact of hyperparameter $r$ in Equation \ref{eq1} by setting the reference value to \emph{background}, \emph{black}, \emph{mean}, and \emph{blur}.  The "black" reference sets $r$ to 0, the "mean" reference sets $r$ to the RGB mean value of the entire dataset, and the "blur" reference applies Gaussian blur (kernel is 39, $\sigma = 15$) to the image and use the blurred pixel value as $r$. Table~\ref{tab:ablation_refv} compares the performance of VisionMask with these four different reference values. 
We choose the background as reference value as it provides a more balanced performance across all metrics. Details about the background reference value are provided in Appendix~\ref{app:sec_ref_value}.

\textbf{Loss Function Components}.
In Table~\ref{tab:ablation_component}, we evaluate the impact of different components in the loss function on the performance of VisionMask. The results show that, although adding negative entropy and L1 area regularization slightly reduces accuracy, it improves the insertion and deletion scores and significantly boosts the Sp./LLE ratio. In contrast, adding total variation to produce smoother masks has minimal impact on overall performance.

\section{Conclusion}

We presented \emph{VisionMask}, an agent-agnostic DRL explanation model trained in self-supervised contrastive learning. 
\emph{VisionMask} generates explanations with higher fidelity and better effectiveness compared to existing attribute-based methods.
It is our future plan to extend this approach to multi-modality input and couple the visual explanation generated by the VisionMask with other information such as agent's long-term goals and future rewards. 

\clearpage
\newpage

\section*{Acknowledgments}
This research is partially supported by the Air Force Office of Scientific Research (AFOSR), under contract FA9550-24-1-0078, and NSF award CNS-2148253.
The paper was received and approved for public release by Air Force Research Laboratory (AFRL) on May 28th 2024, case number  AFRL-2024-2908 . Any opinions, findings, and conclusions or recommendations expressed in this material are those of the authors and do not necessarily reflect the views of AFRL or its contractors.

\bibliographystyle{named}
\bibliography{ijcai25}

\clearpage
\newpage
\appendix
\section{Agent details} \label{app:agent}
\begin{table}[htbp]
\centering
\begin{tabular}{ll}
config & value \\
\toprule
learning rate & 0.001 \\
batch size & 32  \\
$\epsilon$ & 0.5  \\
$\gamma$ & 0.99 \\
buffer size & 50000 \\
target update interval & 1000\\
\end{tabular}
\caption{\textit{VizDoom} DQN}
\label{tab:app_dqn_vizdoom}
\end{table}
\begin{table}[htbp]
\centering
\begin{tabular}{ll}
config & value \\
\toprule
learning rate & 0.0005 \\
batch size & 32  \\
$\epsilon$ & 0.5  \\
$\gamma$ & 0.8 \\
buffer size & 15000 \\
target update interval & 50\\
\end{tabular}
\caption{\textit{Highway-env} DQN}
\label{tab:app_dqn_highway}
\end{table}

As mentioned in the section~\ref{sec_exp}, we adopt the pretrained PPO models from stable-baselines3 \cite{stable-baselines3} for \textit{Enduro, Seaquest} and \textit{MsPacman}. Check the Hugging Face repositories for detailed information on the PPO agents for \textit{Enduro}\footnote{\url{https://huggingface.co/sb3/ppo-EnduroNoFrameskip-v4}}, \textit{Seaquest}\footnote{\url{https://huggingface.co/sb3/ppo-SeaquestNoFrameskip-v4}}, and \textit{MsPacman}\footnote{\url{https://huggingface.co/sb3/ppo-MsPacmanNoFrameskip-v4}}. For \textit{SMB}, we use SMB-PPO\cite{super-mario-bros-PPO-pytorch}\footnote{\url{https://github.com/vietnh1009/Super-mario-bros-PPO-pytorch}}. Inputs for above agents are 4 continuous RGB frames without frame skipping of shape $84 \times 84$.
For \textit{VizDoom}, we train the DQN agent from scratch. The agent consists of four convolutional layers, each with a kernel size of 3, stride of 1, and filter sizes ranging from 64 to 512 (64, 128, 256, 512). Inputs are 5 greyscale frames wihout frame skipping of shape $60 \times 40$. The training setting is in Table~\ref{tab:app_dqn_vizdoom}.
For \textit{Highway-env}, we train the DQN agent of 2 convolutional layers (kernel size 3, stride 1 and filter size 32). Inputs are 4 greyscale frames without frame skipping of shape $128 \times 64$. Details of training is in Table~\ref{tab:app_dqn_highway}. 

\section{Datasets} \label{app:dataset}
\begin{table}[htbp]
\centering
\resizebox{\linewidth}{!}{
\begin{tabular}{lllllll}
env & train & val & test & state dim & state format & actions \\
\toprule
\textit{SMB} & 7754 & 970 & 969 & $4\times256\times240$ & RGB & 7 \\
\textit{Enduro} & 8000 & 1000 & 1000 & $4\times160\times210$ & RGB & 9 \\
\textit{Seaquest} & 8000 & 1000 & 1000 & $4\times160\times210$ & RGB & 9 \\
\textit{MsPacman} & 8000 & 1000 & 1000 & $4\times160\times210$ & RGB & 9 \\
\textit{VizDoom} & 13336 & 1667 & 1667 & $5\times60\times40$ & Grey & 7 \\
\textit{Highway-env} & 15940 & 1992 & 1993 & $4\times64\times128$ & Grey & 5 \\
\end{tabular}
}
\caption{Datasets details of six environments}
\label{tab:app_dataset}
\end{table}
With the aforementioned agents, we collect total state-action pairs of 9693 for \textit{SMB} in 31 different worlds and stages following~\cite{super-mario-bros-PPO-pytorch}, 10000 for \textit{Enduro, Seaquest} and \textit{MsPacman}, 16670 for \textit{VizDoom} and 19925 for \textit{Highway-env}. See Table~\ref{tab:app_dataset} for details.
The datasets is publicly available~\footnote{\url{https://huggingface.co/datasets/0pow0/visionmask_datasets}}

\begin{table*}[h]
    \centering
    \begin{tabular}{llll}
        \toprule
        Mario &  Meaning & Enduro &  Meaning \\
        \midrule
        0 & Stay in place & 0 & No operation \\
        1 & Move right & 1 & Accelerate \\
        2 & Jump right & 2 & Move right \\
        3 & Move right with acceleration & 3 & Move left \\
        4 & Jump right with acceleration & 4 & Decelerate \\
        5 & Jump & 5 & Move right and decelerate \\
        6 & Move left & 6 & Move left and decelerate \\
          & & 7 & Accelerate and move right \\
          & & 8 & Accelerate and move left \\ 
        \bottomrule
        Seaquest &  Meaning & MsPacman &  Meaning \\
        \midrule
        0 & No operation & 0 & No operation \\
        1 & Fire & 1 & Move up\\
        2 & Move up & 2 & Move right \\
        3 & Move right & 3 & Move left \\
        4 & Move left & 4 & Move down \\
        5 & Move down & 5 & Move up-right \\
        6 & Move up-right & 6 & Move up-left \\
        7 & Move up-left & 7 & Move down-right \\
        8 & Move down-right & 8 & Move down-left \\
        9 & Move down-left & & \\
        10 & Fire and move up & & \\
        11 & Fire and move right & & \\
        12 & Fire and move left & & \\
        13 & Fire and move down & & \\
        14 & Fire and move up-right & & \\
        15 & Fire and move up-left & & \\
        16 & Fire and move down-right & & \\
        17 & Fire and move down-left & & \\
        \bottomrule
        VizDoom &  Meaning & Highway-env &  Meaning \\
        \midrule
        0 & TURN LEFT & 0 & LANE\ LEFT \\
        1 & TURN RIGHT & 1 & IDLE \\
        2 & ATTACK & 2 & LANE\ RIGHT \\
        3 & MOVE\ LEFT & 3 & FASTER \\
        4 & MOVE\ RIGHT & 4 & SLOWER \\
        5 & MOVE\ FORWARD & & \\
        6 & MOVE\ BACKWARD & & \\
    \end{tabular}
    \caption{Action meanings for \textit{SMB},  \textit{Enduro}, \textit{Seaquest} and \textit{MsPacman}}
    \label{tab:app_actions}
\end{table*}

\section{Actions}\label{app:actions}
In Table \ref{tab:app_actions}, we present the list of actions and their corresponding meanings for the six environments evaluated in the experiment section.

\section{Example of the human studies}\label{app:human_study}
\begin{figure}[htbp]\centering
\includegraphics[width=0.8\linewidth]{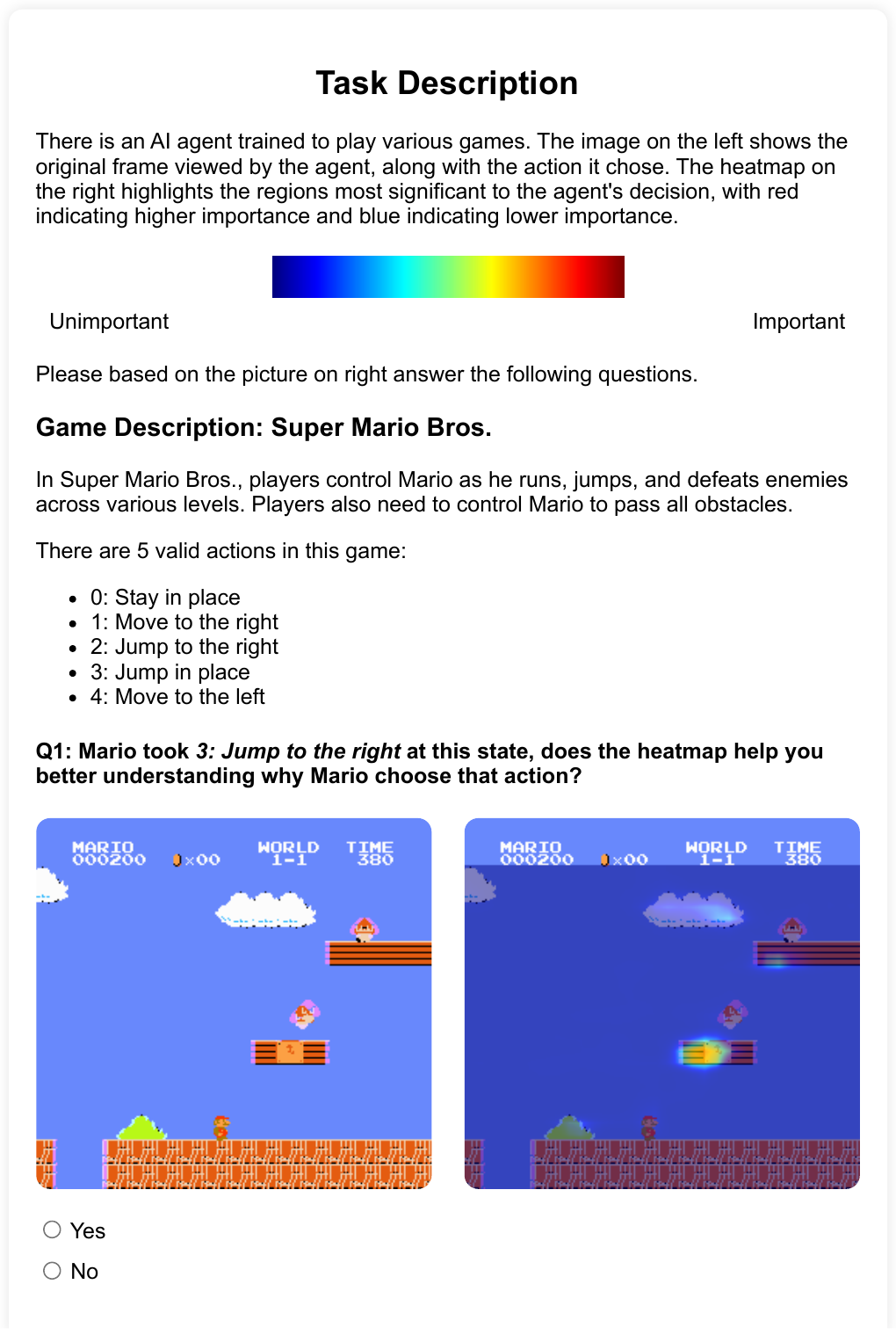}
\caption{Example interface for our human studies}
\label{fig:survey_example}\vspace{-2mm}
\end{figure}
In Figure~\ref{fig:survey_example}, we present a screenshot of the interface used for our human studies. We selected three examples from \textit{SMB}, \textit{Enduro}, and \textit{Seaquest}. Participants were asked to provide Yes/No responses based on whether the saliency map's explanation aligned with their understanding and if it was reasonable.

\section{Algorithms for insertion and deletion metrics} \label{app:alg_del}
Algorithms \ref{alg:ins}-\ref{alg:del} present the pseudocode for evaluating the insertion and deletion metrics of \cite{petsiuk2018rise}.
\begin{algorithm}[H]
    \caption{Evaluate insertion metric}\label{alg:ins}
      \begin{algorithmic}[1]
        \State \textbf{Input:} target mask $m$, state $s$, action $a$, expert policy $\pi_E$, fraction of pixels inserted $\alpha$, Number of pixels $N$
        \State Initialize $m^\prime \gets \text{zeros}(m)$, $p \gets \emptyset$
        \State $\mathcal{I} \gets \text{indices} ( \text{sort} ( \text{flatten} ( m ) ))$ %
        \For{$k \gets 0$ to $\lfloor \alpha N \rfloor -1$}
            \State $(x, y) \leftarrow \mathcal{I}[k]$, $m^\prime[x, y] \gets 1.0$
            \State $p \leftarrow p \cup \pi_E(s \odot m^\prime + r \odot (1 - m^\prime))$
        \EndFor
        \State \textbf{return} $\text{AUC}(p)$
      \end{algorithmic}
    \end{algorithm}

\begin{algorithm}[H]
    \caption{Evaluate deletion metric}\label{alg:del}
    \label{alg:del}
      \begin{algorithmic}[1]
        \State \textbf{Input:} target mask $m$, state $s$, action $a$, expert policy $\pi_E$, fraction of pixels deleted $\alpha$, Number of pixels $N$
        \State \textbf{Initialize} $m^\prime \gets \text{ones}(m)$, $p \gets \emptyset$
        \State $\mathcal{I} \leftarrow \text{indices} ( \text{sort} ( \text{flatten} ( m ) ))$ %
        \For{$k \gets 0$ to $\lfloor \alpha N \rfloor -1$}
            \State $(x, y) \leftarrow \mathcal{I}[k]$, $m^\prime[x, y] \gets 0.0$
            \State $p \leftarrow p \cup \pi_E(s \odot m^\prime + r \odot (1 - m^\prime))$
        \EndFor
        \State \textbf{return} $\text{AUC}(p)$
      \end{algorithmic}
\end{algorithm}

\begin{table*}[htbp]
\captionof{figure}{Reference values for four environment.
The content column represents a sub-figure of the original state that contains the objects and background. The last column displays the corresponding reference value for each environment.}
\label{ref_value}
\small
\centering
\begin{tabular}{cccc}
    \toprule
    Env & Original state & Content & Reference value\\
    \midrule
    SMB &
        \includegraphics[width=0.22\textwidth]{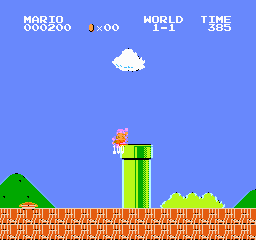} &
        \includegraphics[width=0.22\textwidth]{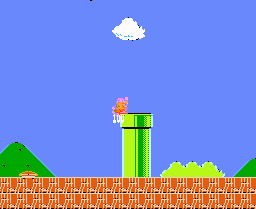} &
        \includegraphics[width=0.22\textwidth]{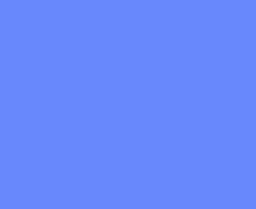} \\
    \midrule 
    Enduro &
        \includegraphics[width=0.22\textwidth]{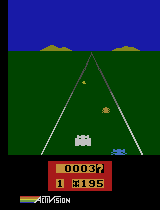} &
        \includegraphics[width=0.22\textwidth]{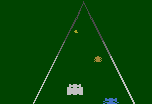} &
        \includegraphics[width=0.22\textwidth]{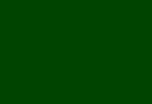} \\
    \midrule
    Seaquest &
        \includegraphics[width=0.22\textwidth]{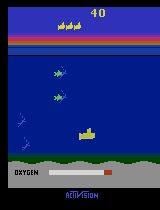} &
        \includegraphics[width=0.22\textwidth]{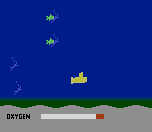} &
        \includegraphics[width=0.22\textwidth]{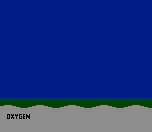} \\
    \midrule
    {MsPacman} &
        \includegraphics[width=0.22\textwidth]{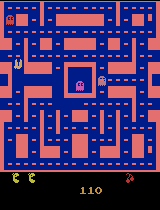} &
        \includegraphics[width=0.22\textwidth]{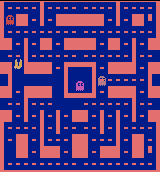} & 
        \includegraphics[width=0.22\textwidth]{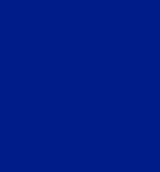} \\
    \bottomrule
\end{tabular}
\end{table*}

\section{Background reference values used across different environments} \label{app:sec_ref_value}
In Figure \ref{ref_value}, we list the reference values we used in each environment.
Specifically, for \textit{SMB} and \textit{Enduro}, we use the background color of the content in each state as the reference value.
For \textit{Seaquest} environment, since all the states share the same background, we use a single reference value for each state, which combines the empty sea background and the oxygen bar background.
For \textit{MsPacman}, we simply used a constant value of the background color blue as the reference value. For \textit{VizDoom}, we use black background; And we use the default grey background for \textit{Highway-env}

\section{Radar map for quantitative results}\label{app:radar}
\begin{figure}[htbp]\centering
\includegraphics[width=0.8\linewidth]{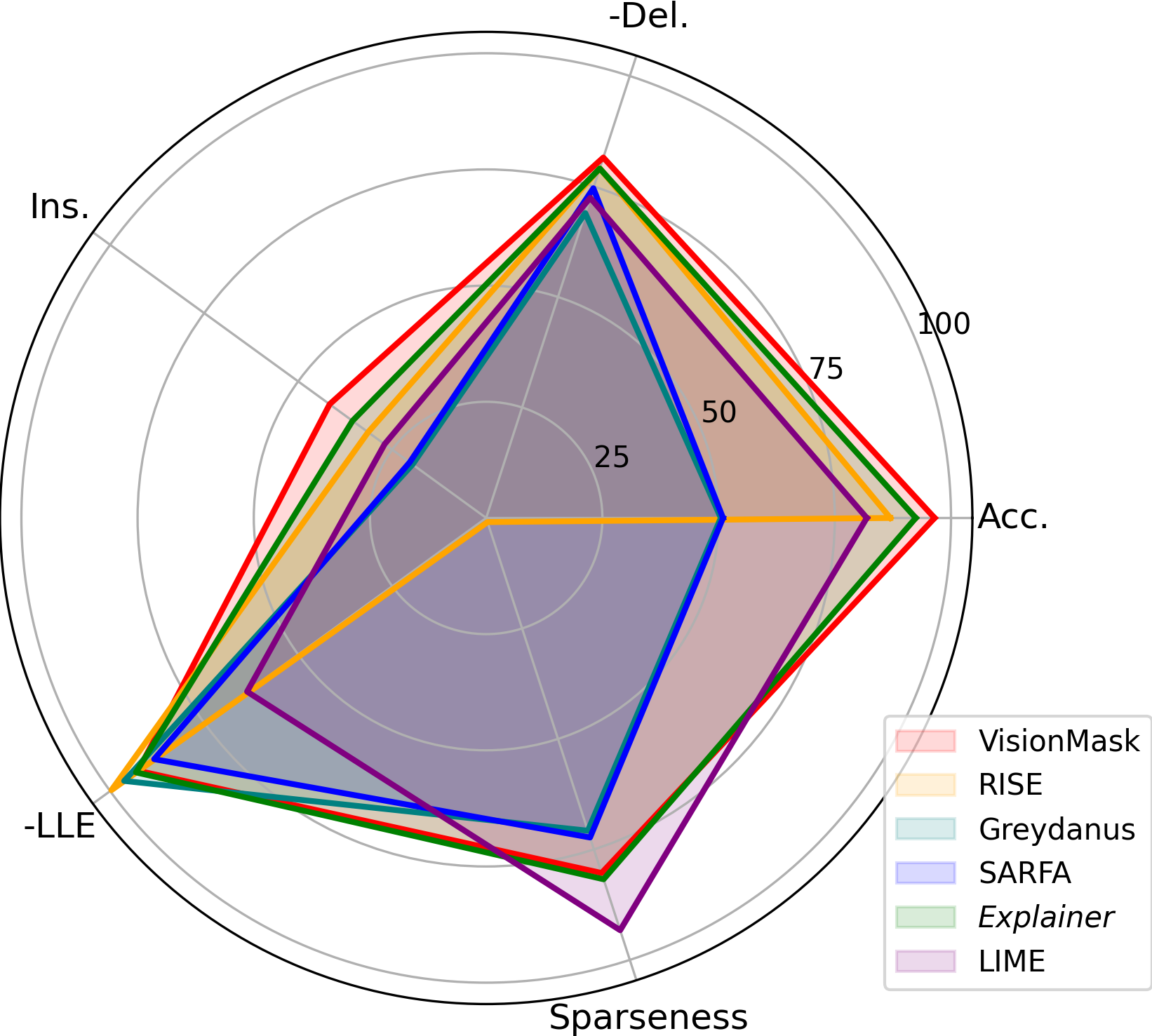}\vspace{-.5em}
\caption{\textbf{Radar map of VisionMask} and five baselines in the six environments. }
\label{fig:radar}\vspace{-2mm}
\end{figure}
We present the radar map in Figure~\ref{fig:radar} to better visualize the quantitative results. Each metric was averaged across all environments. For deletion (Del.) and LLE, we calculated the complement by subtracting the value from 100\%. VisionMask demonstrates better faithfulness (shown in red) compared to other baselines.

\begin{figure*}[t]
\vspace{-.2em}
\centering
\captionsetup[subfloat]{labelformat=empty,font=footnotesize}
\captionsetup[table*]{font=footnotesize}
\subfloat[
]{
\centering
\begin{minipage}{\linewidth}{\begin{center}
\includegraphics[width=0.98\linewidth]{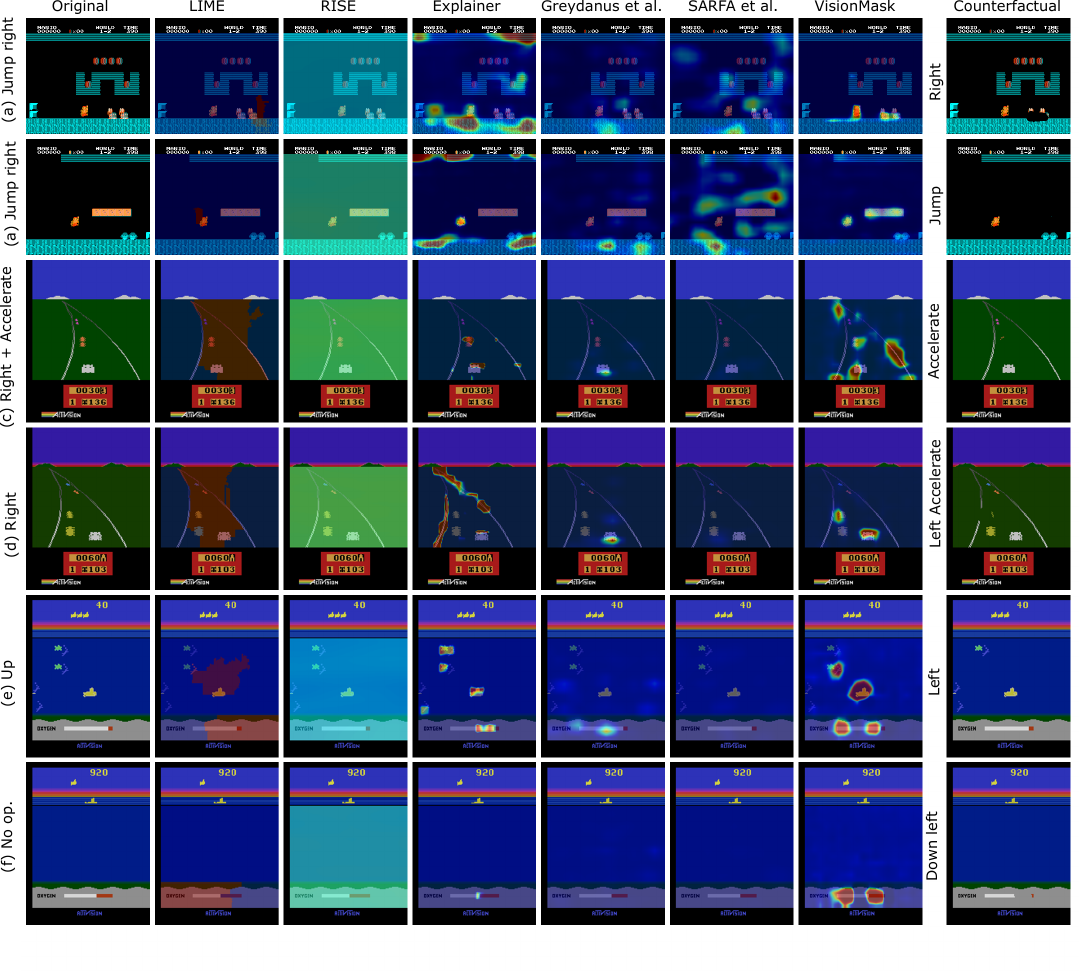}
\end{center}}\end{minipage}
}
\vspace{-2em}
\caption{
\textbf{Additional qualitative examples} compared with five baselines in three environments.
(a) Human explanation: "Mario jumps to the right to get the Koopa on the ground." VisionMask correctly identified the Koopa. Counterfactual analysis shows that removing the two Koopas on the ground changes the action from 'Jump right' to 'Right.' Note that 'Right' cannot reach the Koopa.
(b) Human explanation: "Mario jumps to the right to land on the platform." VisionMask correctly identified the platform. Counterfactual analysis shows that removing the platform changes the action from 'Jump right' to 'Jump.'
(c) Human explanation: "The agent moves to the right while maintaining its current speed because it is blocked by the car in front." VisionMask correctly identified the agent's current location, the front car, and the track. Counterfactual analysis shows that removing the front cars allows the agent to accelerate instead of choosing 'Right + accelerate.'
(d) Human explanation: "The agent moves to the right to avoid a crash with the car on the left." VisionMask correctly identified the agent's current location and the two cars on the left. Counterfactual analysis shows that removing the cars on the left enables the agent to choose 'Left + accelerate' instead of 'Right + accelerate.'
(e) Human explanation: "The agent moves up to rescue the diver." VisionMask correctly identified the relative locations of the agent and the diver, as well as the oxygen bar. Counterfactual analysis shows that removing the diver causes the submarine to stop moving up.
(f) Human explanation: "The agent stays on the surface to wait for the oxygen to fully recharge." VisionMask correctly identified the oxygen bar. Counterfactual analysis shows that removing the rightmost oxygen bar causes the submarine to start operating.
}
\label{fig:app_quali_and_cf} \vspace{-1em}
\end{figure*}

\section{Additional qualitative explames} \label{app:additional_quali}
We present further qualitative examples comparing VisionMask to baselines in \textit{SMB}, \textit{Enduro} and \textit{Seaquest} environments in Figure~\ref{fig:app_quali_and_cf}.

\section{Object importance percentage calculation}
\label{app:obj_importance}

Attention percentages paid to objects are calculated by summarizing the attention map on target objects. Existing object detection models, trained on real-world images, perform unsatisfactorily in gaming environments. Furthermore, most object detection models use bounding boxes to detect objects, which are not suitable for fine-grained attention percentage calculations. To address these challenges, we use a segmentation model as an alternative approach. Specifically, we employ Segment Anything (SAM)~\cite{kirillov2023segany}, which automatically generates masks indicating potential target objects, denoted as $\mathbf{m}^\text{SAM}\in{\{0,1\}^{N^{\text{SAM}}\times{W}\times{H}}}$, where $N^{\text{SAM}}$ is the number of potential target objects. We then multiply the vision mask $m_i\in{\{0,1\}^{W\times{H}}}$ of the ground truth action $i$ element-wise with each target object mask $m^\text{SAM}_i\in{\{0,1\}^{W\times{H}}}$ to obtain attention percentages $\mathbf{p}\in{\mathbb{R}}$ for all potential target objects. By re-ranking $\mathbf{p}$ and filtering out values below a predefined threshold, we derive the attention percentages on target objects. Finally, we manually label the target objects. Note that we preprocess the vision mask $m_i$ for attention percentage calculation by setting values below the average to zero. This preprocessing step filters out objects that received less attention but still show high attention percentages due to their large size, such as the background sky. The hyper-parameters used in SAM automatic mask generator: $\text{points\_per\_side}=64, \text{pred\_iou\_thresh}=0.7, \text{stability\_score\_thresh}=0.8$. The threshold we set on attention percentage calculation is $0.05$.

\begin{table*}[htbp]
\caption{Counterfactual explanations of SMB environment.}
\label{tab:app_mario}
\small
\centering
\begin{tabular}{cccc}
    \toprule
    Original state & VisionMask & \multicolumn{2}{|c}{Counterfactuals} \\
    \midrule
    {
            \includegraphics[width=0.22\textwidth]{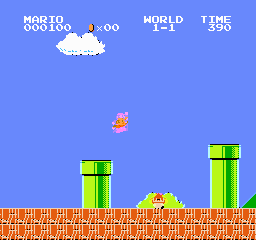}
    } &
 {
        \includegraphics[width=0.22\textwidth]{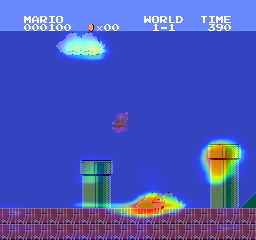}
 } &
 {
        \includegraphics[width=0.22\textwidth]{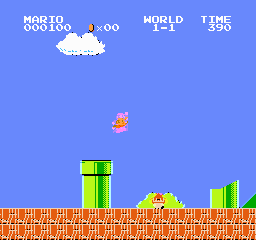}
 } &
 {
        \includegraphics[width=0.22\textwidth]{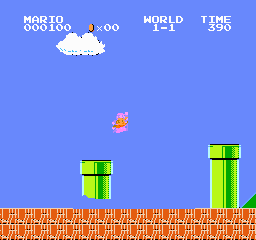}
 }
 \\
 \multicolumn{2}{c}{Action 4}  & Action 5  & Action 5\\
 \multicolumn{2}{c}{Jump right with acceleration}  & Jump & Jump\\
 \midrule
  {
        \includegraphics[width=0.22\textwidth]{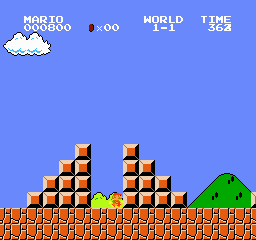}
  }&
  {
        \includegraphics[width=0.22\textwidth]{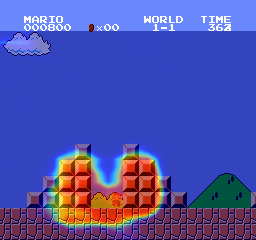}
  }&
  \multicolumn{2}{l}{
        \includegraphics[width=0.22\textwidth]{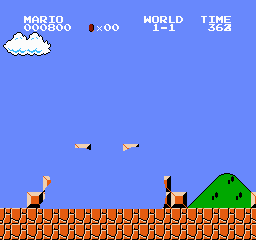}
  }
  \\
  \multicolumn{2}{c}{Action 4} & \multicolumn{2}{l}{Action 3}\\
  \multicolumn{2}{c}{Jump right with acceleration} & \multicolumn{2}{l}{Move right with acceleration}\\
  \midrule
  {
        \includegraphics[width=0.22\textwidth]{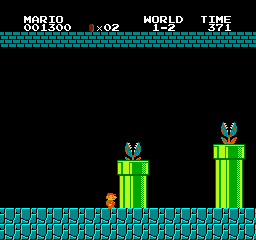}
  }&
  {
        \includegraphics[width=0.22\textwidth]{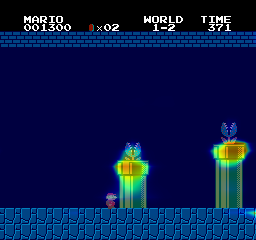}
  }&
  {
        \includegraphics[width=0.22\textwidth]{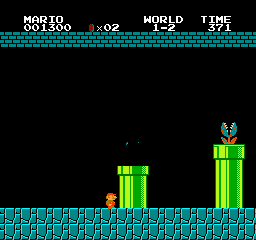}
  }&
  {
        \includegraphics[width=0.22\textwidth]{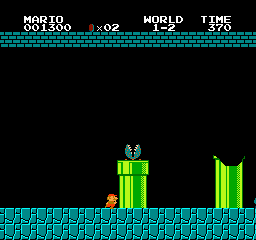}
  }
  \\
  \multicolumn{2}{c}{Action 6} & Action 4 & Action 1\\
  \multicolumn{2}{c}{Move left} & Jump right with acceleration & Move right\\
  \midrule 
  {
        \includegraphics[width=0.22\textwidth]{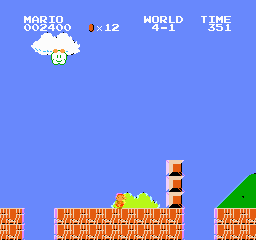}
  } &
  {
        \includegraphics[width=0.22\textwidth]{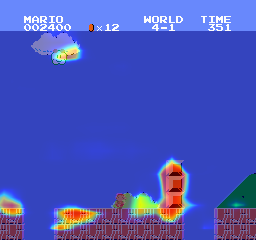}
  } &
  {
        \includegraphics[width=0.22\textwidth]{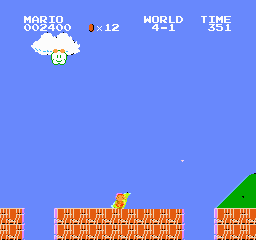}
  } &
  {
        \includegraphics[width=0.22\textwidth]{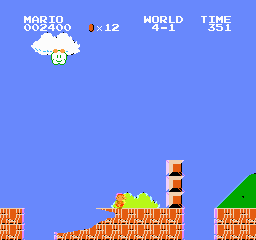}
  } \\
  \multicolumn{2}{c}{Action 2} & Action 4 & Action 2\\
  \multicolumn{2}{c}{Jump right} & Jump right with acceleration & Jump right\\
  \midrule 
\end{tabular}
\end{table*}

\section{Code}
\label{app:code}
We have made our trained models and testing script publicly available. The code can be accessed at \url{https://anonymous.4open.science/r/I6eV3cksE5AgZ7-EB88}. The full set of code will be published in the camera-ready version.

\section{Counterfactual explanations of Super Mario Bros environment}
\label{app:sec_cf_mario}

In this section, we show more counterfactual explanations on the Super Mario Bros\cite{gym-super-mario-bros} environment.
In Table \ref{tab:app_mario} below, each row represents a random state from the environment. The original state column shows the state, the VisionMask column displays the importance map of the target action (label), and the Counterfactuals column presents the corresponding counterfactuals constructed by removing the most important regions indicated by the importance map.
We detail the counterfactual explanations in Table \ref{tab:app_mario} below:

\begin{itemize}
\item In the first row, Mario encounters the monster Goomba and two pipes, with the bush and Goomba receiving an importance score of 31.62\%, while the right pipe receives 29.2\%.
Counterfactual scenarios involving the removal of the Goomba and the top section of the second green pipe both result in a change of action from accelerating while jumping right to simply jumping,
which may indicate that those regions are the cause of action 4 (jump right with acceleration).

\item In the second row, Mario is positioned between two brown bricks, with the left brick receiving an importance score of 25.23\%, the right brick receiving 29.15\%, and Mario and the bush each receiving 25.23\%.
It's worth noting that jumping is necessary in the original state for Mario to pass these obstacles, but it becomes unnecessary in the counterfactual scenario.
This observation suggests that the bricks may be the cause of Action 4, as when they are removed, jumping is no longer necessary.

\item The third row is identical to the example presented in the counterfactual explanation section of the Super Mario Bros (SMB) environment in the paper.

\item In the fourth column, Mario encounters a brick wall that only requires a jump (without acceleration) to pass through, with the brick wall receiving an importance score of 28.78\%.
The first counterfactual, removing the brick, indicates a change in action from Action 2 (right jump) to Action 4.
Knowing that a right jump with acceleration (Action 4) is necessary to safely jump over the gap and land on the next platform, one can easily identify the brick wall as the cause of the action.
In addition, the observation that the second counterfactual of removing the ground behind Mario did not change the action, which in turn supports our previous explanation.
\end{itemize}

Note that all the object importance percentages are calculated following \ref{app:obj_importance}.

\begin{table*}[htbp]
\caption{Counterfactual explanations of Atari Enduro environment. Note that the collision happens in the original state in the third row.}
\label{tab:app_enduro}
\centering
\begin{tabular} {cccc}
  \toprule
    Original state & VisionMask & \multicolumn{2}{|c}{Counterfactuals} \\
    \midrule
    {
            \includegraphics[width=0.2\textwidth]{
                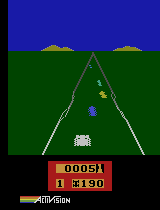
            }
    } &
    {
            \includegraphics[width=0.2\textwidth]{
                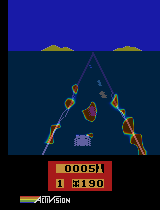
            }
    } &
    \multicolumn{2}{c}{
            \includegraphics[width=0.2\textwidth]{
                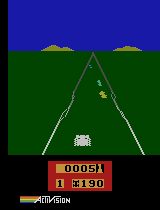
            }
    }
    \\
    \multicolumn{2}{c}{Action 3: left}  & \multicolumn{2}{c}{Action 8: left + accelerate}
    \\
    \midrule
     {
            \includegraphics[width=0.2\textwidth]{
                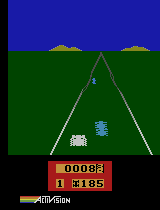
            }
    } &
    {
            \includegraphics[width=0.2\textwidth]{
                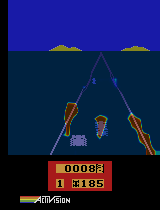
            }
    } &
    \multicolumn{2}{c}{
            \includegraphics[width=0.2\textwidth]{
                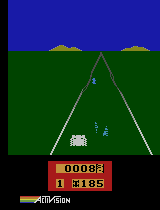
            }
    }
    \\
    \multicolumn{2}{c}{Action 1: accelerate}  & \multicolumn{2}{c}{Action 7: right + accelerate}
    \\
    \midrule
    {
            \includegraphics[width=0.2\textwidth]{
                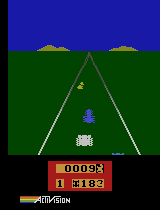
            }
    } &
    {
            \includegraphics[width=0.2\textwidth]{
                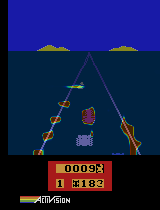
            }
    } &
    \multicolumn{2}{c}{
            \includegraphics[width=0.2\textwidth]{
                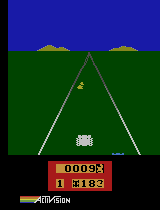
            }
    }
    \\
    \multicolumn{2}{c}{Action 3: left}  & \multicolumn{2}{c}{Action 8: left + accelerate}
    \\
    \midrule
    {
            \includegraphics[width=0.2\textwidth]{
                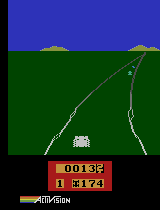
            }
    } &
    {
            \includegraphics[width=0.2\textwidth]{
                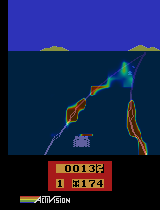
            }
    } &
    \multicolumn{2}{c}{
            \includegraphics[width=0.2\textwidth]{
                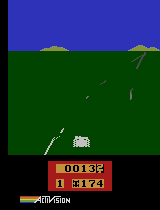
            }
    }
    \\
    \multicolumn{2}{c}{Action 7: right + accelerate}  & \multicolumn{2}{c}{Action 8: left + accelerate}
    \\   
    \bottomrule
\end{tabular}
\end{table*}

\section{Counterfactual explanations of Atari Enduro environment}
\label{app:sec_cf_enduro}

In this section, we presents counterfactual explanations on the Atari Enduro environment.
In Table \ref{tab:app_enduro} below, each row represents a random state from the environment.
The original state column shows the state, the VisionMask column displays the importance map of the target action (label), and the Counterfactuals column presents the corresponding counterfactuals constructed by removing the most important regions indicated by the importance map.
We detailed the counterfactual explanations in Table \ref{tab:app_enduro} below:

\begin{itemize}
\item In the first row, the agent attempts to move left to avoid a collision with the incoming blue car, which receives an importance score of 8.33\% as indicated by the importance map of second column.
The counterfactual scenario involving the removal of the blue car leads to a change in action from moving left to accelerating while moving left.
It's important to note that accelerating while moving left could still result in a collision with the blue car in the original state.

\item In the second row, the agent attempts to overtake the blue car on the right, which receives an importance score of 11.25\%.
The corresponding counterfactual scenario of removing the blue car indicates a transition of action from accelerating to moving right and accelerating.
Given that moving right with accelerating will cause a collision with that car, we may attribute the accelerating action to the blue car.

\item In the third row, a collision occurs in the original state with Action 3 (moving left).
The agent tries to move left to avoid the collision but fails, where the collided blue car receives an importance score of 16.10\%.
By removing the blue (collided) car from the original state, the action in that counterfactual scenario changes to moving left with accelerating
One may explain the collided blue car as the main cause of the action (moving left) in this counterfactual scenario.

\item In the last row, the agent tries to turn right, where the race track receives an importance score of 27.58\%.
In the counterfacutal scenario, the action changes from moving right with acceleration to moving left with acceleration when most of the race track is removed from original state.
It's obvious that the right directioned race track is the main cause of action of truing right with accelerating.
\end{itemize}

\begin{table*}[htbp]
\caption{Counterfactual explanations of Atari Seaquest environment}
\label{tab:app_sea}
\centering
\begin{tabular} {cccc}
  \toprule
    Original state & VisionMask & \multicolumn{2}{|c}{Counterfactuals} \\
    \midrule
    {
            \includegraphics[width=0.2\textwidth]{
                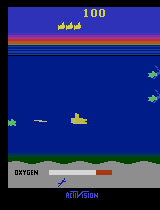
            }
    } &
    {
            \includegraphics[width=0.2\textwidth]{
                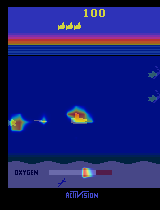
            }
    } &
    \multicolumn{2}{c}{
            \includegraphics[width=0.2\textwidth]{
                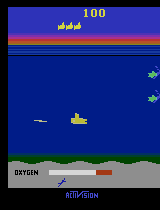
            }
    } 
    \\
    \multicolumn{2}{c}{Action 13: down + fire}  & \multicolumn{2}{c}{Action 8: down + right}
    \\
    \midrule 
    {
            \includegraphics[width=0.2\textwidth]{
                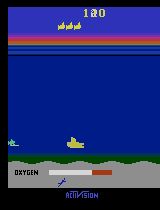
            }
    } &
    {
            \includegraphics[width=0.2\textwidth]{
                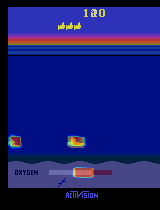
            }
    } &
    \multicolumn{2}{c}{
            \includegraphics[width=0.2\textwidth]{
                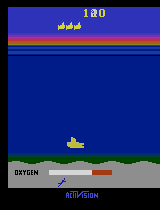
            }
    } 
    \\
    \multicolumn{2}{c}{Action 10: up + fire}  & \multicolumn{2}{c}{Action 6: up + right}
    \\
    \midrule 
    {
            \includegraphics[width=0.2\textwidth]{
                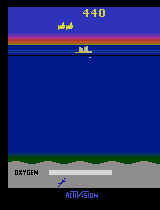
            }
    } &
    {
            \includegraphics[width=0.2\textwidth]{
                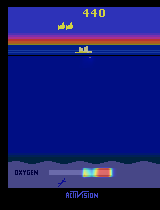
            }
    } & 
    \multicolumn{2}{c}{
            \includegraphics[width=0.2\textwidth]{
                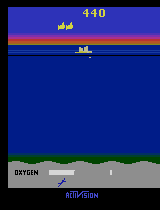
            }
    } 
    \\
    \multicolumn{2}{c}{Action 17: down + left + fire}  & \multicolumn{2}{c}{Action 4: left}
    \\
    \midrule  
\end{tabular}
\end{table*}

\section{Counterfactual explanations of Atari Seaquest environment}
\label{app:sec_cf_sea}

In this section, we presents counterfactual explanations on the Atari Seaquest environment.
In Table \ref{tab:app_sea} below, each row represents a random state from the environment.
The original state column shows the state, the VisionMask column displays the importance map of the target action (label), and the Counterfactuals column presents the corresponding counterfactuals constructed by removing the most important regions indicated by the importance map.
We detailed the counterfactual explanations in Table \ref{tab:app_sea} below:

\begin{itemize}
\item In the first row, the agent attempts to attack the shark in the bottom left corner, marked in green, while the shark receives an importance score of 5.65\%.
In the counterfactual scenario where the shark is removed from the original state, the resulting action shifts from moving down and firing to moving down and to the right.
The absence of the shark reasonably leads to a shift in action from firing to moving right, indicating that the presence of the shark likely prompted the original action.

\item The second row is similar to first row where the counterfactual of removing shark could incur the action shifting from fire to a movement.

\item In the last row, the agent's submarine is about to start diving with a full tank of oxygen, where the oxygen bar receives most of the importance score, 53.51\%.
Counterfactual of removing the half of the oxygen bar making the action change from moving down, left and fire to moving left to stay on the surface.
The oxygen level can easily be explained as the main cause in that counterfactual scenario.
\end{itemize}

\end{document}